\pdfoutput=1

\documentclass[applsci,article,accept,moreauthors,pdftex]{mdpi} 

\firstpage{1} 
\pubvolume{9}
\issuenum{8}
\articlenumber{0}
\pubyear{2019}
\copyrightyear{2019}
\history{Received: 21 March 2019; Accepted: 16 April 2019; Published: date}
\updates{yes} 

\setitemize{parsep=6pt,itemsep=0pt,leftmargin=*,labelsep=5.5mm,align=parleft}
\setenumerate{parsep=6pt,itemsep=0pt,leftmargin=*,labelsep=5.5mm,align=parleft}

\usepackage{listings}
\lstdefinestyle{customc}{
  belowcaptionskip=1\baselineskip,
  breaklines=true,
  xleftmargin=\parindent,
  language=C,
  showstringspaces=false,
  basicstyle=\footnotesize\ttfamily,
  keywordstyle=\bfseries\color{green!40!black},
  commentstyle=\itshape\color{purple!40!black},
  identifierstyle=\color{black},
  stringstyle=\color{orange},
}
\usepackage[caption=false,font=footnotesize,labelformat=simple]{subfig}

\usepackage[ruled,vlined]{algorithm2e}
\newcommand{\mathds}[1]{\mathbb{#1}}
\newcommand{\beq}{\begin{equation}}
\newcommand{\eeq}{\end{equation}}
\newcommand{\R}{\mathds{R}}

\DeclareMathOperator*{\argmin}{arg\,min}

\newcommand{\prox}{\textnormal{prox}}

\usepackage{bbm}

\Title{Compressed Learning of Deep Neural Networks for OpenCL-Capable Embedded Systems}

\Author{Sangkyun Lee~$^{1,}$*\orcidA{} and Jeonghyun Lee~$^{2}$}

\AuthorNames{Sangkyun Lee and Jeonghyun Lee}

\address{%
  $^{1}$ \quad Computer Science, Hanyang University ERICA, Ansan 15588, Korea\\
  $^{2}$ \quad Computer Science and Engineering, Hanyang University ERICA,  Ansan 15588, Korea; nomar0107@hanyang.ac.kr}

\corres{Correspondence: sangkyun@hanyang.ac.kr; Tel.:~+82-31-400-1039}

\abstract{Deep neural networks (DNNs) have been quite successful in
  solving many complex learning problems. However, DNNs tend to have a
  large number of learning parameters, leading to a large memory and
  computation requirement. In this paper, we propose a model
  compression framework for efficient training and inference of deep
  neural networks on embedded systems. Our framework provides data
  structures and kernels for OpenCL-based parallel forward and
  backward computation in a compressed form. In particular, our method
  learns sparse representations of parameters using $\ell_1$-based
  sparse coding while training, storing them in compressed sparse
  matrices. Unlike the previous works, our method does not require a
  pre-trained model as an input and therefore can be more versatile
  for different application environments. Even though the use of
  $\ell_1$-based sparse coding for model compression is not new, we
  show that it can be far more effective than previously reported when
  we use proximal point algorithms and the technique of debiasing. Our
  experiments show that our method can produce minimal learning models
  suitable for small embedded devices.}


\begin{document}

\section{Introduction}

Modern deep neural networks (DNNs) tend to have growing numbers of
parameters, which are often unavoidable to solve complex learning
problems in computer vision~\citep{KriS12}, speech
recognition~\citep{GraS05}, and natural language processing
problems~\citep{ColW11}. For example, the number of parameters of
convolutional neural networks (CNNs) for computer vision tasks has
grown. Compared to the first CNN, Lenet-5~\citep{LecB98}, which
contains less than 1M parameters, recent networks such as
AlexNet~\citep{KriS12} (60M) and VGG16~\citep{SimZ15} (138M) consist
of a much larger number of parameters.

Some networks have been optimized for their size, for example,
GoogleNet (also known as Inception v1)~\citep{SzeL15}, the winning
method of ImageNet Large Scale Visual Recognition Challenge (ILSVRC)
in 2014, only has 6.7M parameters, a surprisingly smaller number
compared to the previous winners (AlexNet~\citep{KriS12} in 2012,
ZFNet~\citep{ZeiF14} in 2013). However, new versions of GoogleNet such
as BN-Inception~\citep{IofS15}, Inception v2 and v3~\cite{SzeV16}, and
Inception v4~\cite{SzeI17} tend to become larger for better prediction
accuracy, for example, Inception v4 contains about 80M
parameters. ResNet-152~\citep{HeZ16}, the winner of ILSVRC 2015
defeating Inception v3, contains 60M parameters. Other examples
include Deepfaces~\citep{TaiY14} (120M parameters) and DNNs on
high-performance computing systems~\citep{CoaH13} (10B parameters).

Large numbers of learning parameters, however, make it preventive to
learn and apply machine learning models on small devices, e.g.,
smartphones, embedded computers, and wearable devices, where memory,
computation, and energy consumption can be restricted. In addition, a large
number of parameters may lead to overfitting, a phenomenon in machine
learning that a complex and large machine learning model fits training
data too much and its generalization performance on future data
decreases as a result~\citep{LawG97}. Therefore, compressing large
machine learning models is an essential consideration for the training
and the deployment of them in application environments such as edge
computing~\citep{BuyY09} with small computation devices.

\subsection{Related Work}

The main idea of model compression is to reduce the effective number
of parameters to be stored and computed. One of the classical ideas
for DNN model compression is to use a low-rank approximation of weight
matrices. This technique has been applied successfully in scene text
character recognition~\citep{JadV14}, and later improved and extended
for larger DNNs~\citep{DenZ14,IoaR16,TaiX16}, reducing memory
consumption and computational cost. However, these approaches work on
fixed network architecture, requiring reiterations of decomposition,
training, fine-tuning, and cross-validation.


\subsubsection{Network Pruning}

For the model compression of DNNs, network pruning is probably the
most popular approach, which tries to remove irrelevant neural
connections associated with weight values below a certain
threshold. In biased weight decay~\citep{HanP89}, hyperbolic and
exponential biases were introduced to the pruning objective, where
brain damage~\citep{LecD90} and brain surgeon~\citep{BabS93} used
information from the Hessian of the loss function.

More recently, \citet{HanP15} suggested a network pruning approach
with retraining, which performs weight training, network pruning, and
then another round of weight training (retraining). The final
retraining is done only for the connections not removed during the
pruning step, in order to eliminate statistical bias caused by network
pruning. The authors reported that pruning with $\ell_2$
regularization and retraining was the most successful regarding
prediction accuracy, compared to $\ell_1$ and $\ell_2$ regularization
alternatives with and without retraining.  This method has further
developed into deep compression~\citep{HanM16}, which adds two extra
steps after pruning: weight quantization and the Huffman
encoding. This method has been followed-up with more emphasis on
compressing weights in convolutional layers, using structured sparsity
regularization to induce structured sparsity in rows, columns,
channels, and filters~\citep{WenW16,LebL16}.

\subsubsection{Network Pruning as Optimization}
Since network pruning can be achieved by adding a regularization term
in the training objective function, model compression by network
pruning can be considered in terms of optimization problems. For
example, stochastic optimization algorithms for $\ell_1$
regularization problems~\citep{SST09,SinD09} have been applied to CNN
model compression~\citep{ZhoA16}. The forward-backward splitting
algorithms in these works are closely related to our proximal point
algorithms~\citep{ParB14} to be discussed later, but only classical
stochastic gradient descent has been considered in the previous work,
whereas we consider more advanced training algorithms for neural
networks such as RMSProp~\citep{Hin14} and ADAM~\citep{KinB15}.

The state-of-the-art network pruning approach~\citep{CarI18} considers
reforming the network pruning into constrained optimization problems,
solving them using the augmented Lagrangian method, also known as the
method of multipliers~\citep{Hes69,Pow69}. Our proposed method also
considers the penalized optimization problem as in this approach, with
an important distinction that we use proximal point
algorithms~\citep{ParB14} which comes with clear benefits: lower
memory consumption and faster convergence, as we show later.

Other works in model compression include guiding compression
with side information such as network latency provided in forms of
optimization constraints~\citep{CheT18}, designing network
architecture using reinforcement learning (e.g., NAS~\citep{ZopL16}
and NASNet~\citep{ZopV17}), and tuning parameters that trade-off model
size and prediction accuracy automatically~\citep{HeL18}. We refer to
\citet{CheW18} for a more detailed survey of DNN model compression.


\subsection{Contribution}
Even though there are quite a few works on model compression for deep
neural networks, we found that they may not be well suited for
learning and inference with DNNs in small embedded~systems:

\begin{itemize}
\item Full model training is required prior to model compression: in
  network pruning~\citep{HanP15,HanM16}, including the
  state-of-the-art method~\citep{CarI18}, and low-rank
  approximation~\citep{DenZ14,IoaR16,JadV14,TaiX16}, full models have
  to be trained first, which would be too burdensome to run on small
  computing devices. In addition, compression must be followed by retraining
  or fine-tuning steps 
  with a substantial number of training
  iterations, since, otherwise, the compressed models tend to show
  impractically low prediction accuracy, as we show in our
  experiments. 

\item Platforms are restricted: existing
  approaches~\citep{HanP15,HanM16,WenW16} have been implemented and
  evaluated mainly on CPUs because sparse weight matrices produced by
  model compression typically have irregular nonzero patterns and
  therefore are not suitable for GPU computation. \mbox{In~\citet{WenW16}},
  some evaluations have been performed on GPUs, but it was implemented
  with a proprietary sparse matrix library (cuSPARSE) only available
  on the NVIDIA platforms (Santa Clara, CA, USA). \citet{HeL18} considered auto-tuning of
  model compression experimented on NVIDIA GPU and a mobile CPU on
  Google Pixel-1 (Mountain View, CA, USA), yet without discussion on how to implement them
  efficiently on embedded~GPUs.
\end{itemize}

In this paper, we provide a model compression method for deep neural
networks based on $\ell_1$-regularized optimization, often referred to
as sparse coding~\citep{LeeB07,NeeT09}. Application of $\ell_1$-based
regularization for DNNs is not entirely new; however, we discuss that
it can be far more effective than previously reported. In particular,
we apply $\ell_1$-based sparse coding with (i) {\em proximal
  operators}, which induce explicit sparsity in learning weights
during training. In addition, it is followed by a (ii) {\em debiasing} step,
which tries to reduce estimation bias due to sparse coding: we discuss
that this technique allows us to compress DNNs further without
sacrificing prediction accuracy much. While inducing sparsity on
learning weights, we also store sparse weights in (iii) {\em
  compressed data structures}, being able to compute forward and
backward passes with compressed weights on any (iv) {\em
  OpenCL-capable devices}, e.g., mobile GPUs such as ARM Mali-T880 (Cambridge, UK).


\section{Method}\label{sec:method}

In this section, we describe the details of our method in which
network pruning is considered as an $\ell_1$-regularized optimization
problem, i.e, sparse coding, where we aim to find a sparse
representation of observations, sometimes also looking for good basis
functions for such a representation (this particular task is called
dictionary learning~\citep{BaoJ16}). The idea of sparse coding has
been applied in various fields of research, i.e., sparse regression in
statistics (e.g., \cite{Tib96}), compressed sensing in signal
processing (e.g.,~\cite{CanT07,CanT05,Don06}), matrix completion in
machine learning (e.g., \cite{CanP10}), just to name a few.

In our approach, we apply sparse coding for model compression, to find
a {\em sparse} optimal weight parameters containing as many zero
values as possible whenever the associated neural connections in a
deep neural network are irrelevant for making an accurate prediction.

\subsection{Sparse Coding in Training}

To describe our method formally, let us suppose that we train a deep
neural network with $n$ examples, $\{(x_i,y_i)\}_{i=1}^n$, where
$x_i \in \R^d$ is an input vector and $y_i \in \R$ (regression) or
$y_i \in \{1,2,\dots,K\}$ (classification) is the corresponding
outcome. In training, we obtain the optimal weight $w^* \in \R^p$ by
solving the following optimization problem:
$$
 w^* \in \argmin_{w\in\R^p} \;\; \frac1n \sum_{i=1}^n \ell(w; x_i, y_i). \enspace 
$$

Here, $\ell(w;x_i,y_i)$ is the loss function describing the gap
between the prediction made by the neural net and the desired
outcome. In sparse coding, we add a regularizer $\Psi(w)$ to the
objective function and solve a modified training problem:
\begin{equation}\label{eq:min}
 w^* \in \argmin_{w\in\R^p} \;\; \frac1n \sum_{i=1}^n \ell(w; x_i, y_i) + \Psi(w). \enspace
\end{equation}

We use the $\ell_1$-norm $\Psi(w) = \lambda \|w\|_1$ for our method in
particular, where $\lambda>0$ is a hyperparameter that controls the
rate of model compression (the larger $\lambda$ is, the resulting
model will be more compressed as more weight values will set to the
zero value).

Note that the use of $\ell_1$-regularizer by itself does not
automatically lead to sparse coding. We need an explicit mechanism
that sets weight values to the zero value while solving
Equation~\eqref{eq:min}. The mechanism is called a proximal operator.

\subsection{Proximal Operators}\label{sec:proximal}

When we solve the training problem \eqref{eq:min} for a deep neural
network with a large dataset, the most popular optimization algorithms
include stochastic subgradient descent~\citep{NemJ09} and its variants
such as AdaGrad~\citep{Duc11}, RMSProp~\citep{Hin14}, and
ADAM~\citep{KinB15}. These methods use small subsamples of training
examples $B \subset \{1,\dots,n\}$, called minibatches, to construct
stochastic subgradients in forms of:
$$
  g(w) := g_B(w) + \nabla \Psi(w), \enspace
$$ where
$ g_B(w) := \frac{1}{|B|} \sum_{i \in B} \nabla \ell(w;x_i,y_i) $ and
$\nabla \Psi(w)$ is a subgradient of $\Psi(w)$. If we use the
subgradient $g(w)$ which includes the part $\nabla \Psi(w)$ from the
regularizer for updating learning parameters $w$, it is unlikely that
any updated weight value will be precisely the zero value, even though
some values might be close to zero.

Instead, we consider a mechanism to set the weight values to the exact
zero value whenever necessary, by means of the so-called proximal
operator of a convex function $\Psi(\cdot)$. Given a vector
$z \in \R^p$, the proximal operator of $z$ with respect to $\Psi$ is
defined as
$$
 \text{prox}_\Psi(z) := \argmin_{w \in \R^p} \; \frac12 \|w-z\|_2^2 + \Psi(w). \enspace 
$$

When $\Psi(w) = \lambda \|w\|_1$, the regularizer we use for sparse
coding, the proximal operator can be computed in a closed form
independently for each component,
$$
  [\text{prox}_\Psi(z)]_i = \text{sgn}(z_i) \max\{|z_i|-\lambda, 0\}, \;\; i=1,\dots, p. \enspace
$$

This operator is also known as the soft-thresholding
operator~\citep{LeeW11b}.


\subsection{Optimization Algorithm}

When the number of training examples $n$ is not too large, say
$n \sim 10k$, we can solve the training problem with sparse
coding~\eqref{eq:min} using proximal gradient
descent~\citep{AusT06,ParB14}, which updates the weight vector at the
$k$th iteration by:
$$
 w^{k+1} \leftarrow \prox_{\eta \Psi} \left(w^k - \eta G(w^k) \right). \enspace
$$

Here, $\eta >0$ is the learning rate and $G(w^k)$ is the full
gradient involving all training examples,
$$
 G(w^k) := \frac1n \sum_{i=1}^n \nabla \ell(w^k; x_i, y_i). \enspace
$$

The proximal gradient descent algorithm is capable of solving the
regularized training problem~\eqref{eq:min} with
$\Psi(w) = \lambda \|w\|_1$, while optimally fixing irrelevant weights
in $w \in \R^p$ to the exact zero value. However, it will be too
costly to use the method for large $n$, since it will require $n$
back-propagation steps to compute the $\nabla \ell(w^k; x_i,y_i)$ for
all training examples in each update.

We suggest to use a minibatch gradient $g_B(w)$ in the proximal
operator so that the update will be:
\begin{equation}\label{eq:prox}
  w^{k+1} \leftarrow \prox_{\eta \Psi} \left(w^k - \eta g_B(w^k) \right). \enspace 
\end{equation}

Due to the stochasticity in minibatch gradients, the behavior of the
proximal gradient algorithm based on the update~\eqref{eq:prox} will
become less predictable, but some convergence results are available
for the algorithm~\citep{Nit14,PatN18,RosV14} (under the assumption
that the loss function $\ell$ is convex and Lipschitz continuous in
$w$, the objective function value converges to the optimal value with
the rate $\mathcal O(k^{-1/2})$ in expectation).

We further consider the update~\eqref{eq:prox} within the
RMSProp~\citep{Hin14} and ADAM~\citep{KinB15} algorithms, arguably the
two most popular methods for training deep neural networks. RMSProp
uses adaptive learning rates computed differently on gradient
components, while ADAM combines the idea of adaptive learning rates
and that of the momentum method~\citep{Pol03} to improve convergence
and to escape saddle points. We integrate proximal operators with
RMSProp and ADAM: the resulting algorithms are called Prox-RMSProp
(Algorithm~\ref{alg:prox-rmsprop}) and Prox-ADAM
(Algorithm~\ref{alg:prox-adam}).  We can conjecture that the search
directions produced by Prox-ADAM will be more stable than those of
Prox-RMSProp since the former uses compositions of minibatch gradients
and momentum directions, not just noisy minibatch gradients as in the
latter.  We show in experiments that the behavior of Prox-ADAM is
indeed more stable than that of Prox-RMSProp.

\begin{figure}[H]
\begin{minipage}{\columnwidth}
\SetAlgoCaptionSeparator{.}
\begin{algorithm}[H]
  \caption{Prox-RMSProp Algorithm}\label{alg:prox-rmsprop}
 \KwIn{a learning rate $\eta$, a safeguard parameter $\epsilon>0$, $\beta \in [0,1)$: a decay rate}
 Initialize the weight vector $w\in \R^p$\;
  $v \leftarrow 0$ (Init. 1st moment vector)\;
  $t \leftarrow 0$ (Init. timestep)\;
 \While{$w_t$ not converged}{
     $t \leftarrow t+1$\;
     Compute a minibatch gradient: $g_t \leftarrow g_B(w_{t-1})$\;
     Update grad info: $v_t \leftarrow \beta v_{t-1} + (1-\beta) g_t \odot g_t$\;
     Update: $w_t \leftarrow \text{prox}_{\eta \lambda \|\cdot\|_1} (w_{t-1} - \eta \cdot g_t / (\sqrt{v_t}+\epsilon))$\;
   }
   \Return{$w_t$}\;
 \end{algorithm}
\end{minipage}
\vspace{12pt}

\begin{minipage}{\columnwidth}
\SetAlgoCaptionSeparator{.}
\begin{algorithm}[H]
  \caption{Prox-ADAM Algorithm}\label{alg:prox-adam}
\KwIn{a learning rate $\eta$, a safeguard parameter $\epsilon>0$, $\beta_1,\beta_2 \in [0,1)$: decay rates}
  Initialize the weight vector $w\in \R^p$\;
  $m_0 \leftarrow 0$ (Init. 1st moment vector)\;
  $v_0 \leftarrow 0$ (Init. 2nd moment vector)\;
  $t \leftarrow 0$ (Init. timestep)\;
 \While{$w_t$ not converged}{
   $t \leftarrow t+1$\;
   Compute a minibatch gradient: $g_t \leftarrow g_B(w_{t-1})$\;
   Update 1st moment: $m_t \leftarrow \beta_1 m_{t-1} + (1-\beta_1) g_t$\;
   Update 2nd moment: $v_t \leftarrow \beta_2 v_{t-1} + (1-\beta_2) g_t \odot g_t$\;
   Bias correction 1st: $\hat m_t \leftarrow m_t / (1- \text{pow}(\beta_1,t))$\;
   Bias correction 2nd: $\hat v_t \leftarrow v_t / (1- \text{pow}(\beta_2,t))$\;
   Update: $w_t \leftarrow \text{prox}_{\eta \lambda \|\cdot\|_1} (w_{t-1} - \eta \cdot \hat m_t / (\sqrt{\hat v_t}+\epsilon))$\;
\Return{$w_t$}\;
}
\end{algorithm}
\end{minipage}
\end{figure}

\subsection{Retraining}

Once we have obtained a sparse model via sparse coding, we can
consider an optional step to train the weights again without any
regularization, starting from the previously trained weight values,
while excluding the zero-valued weights from training. This type of
retraining is known as {\em debiasing}~\citep{WriNF09}, which can be
used to remove the unwanted reduction of weight values due to
regularization. It has been shown that debiasing can improve the
quality of estimation~\citep{FigN07}, although it may undo some
desired effect of regularization, e.g., the reduction of distortion
caused by noise~\citep{Don95}. Our methods work well without
retraining, but our experiments show that retraining can achieve
further compression without sacrificing prediction accuracy.

\section{Accelerated OpenCL Operations}

In order to perform forward and backward computations of deep neural
networks in an accelerated manner, we must use an efficient
representation of sparse weight matrices, which can be also stored in
a compact form to reduce memory footprint. In the previous research,
sparse data structure and matrix operations have not been discussed
fully enough~\citep{HanP15,HanM16,WenW16}, due to the difficulty of
using sparse matrix operations efficiently on GPUs. Therefore,
implementations and testing were performed mainly on
CPUs~\citep{HanP15,HanM16}, or some vendor-specific support was used,
e.g., the cuSPARSE library, which is not an open-source software and
available only for NVIDIA hardware~\citep{WenW16}.

In this section, we discuss the details how we implement sparse weight
matrices and efficient sparse matrix operations for training deep
neural networks with sparse coding, making use of our proposed
algorithms. We base our discussion on the
OpenCL-Caffe~\citep{oclcaffe} 
software, an OpenCL-capable version of the favorite open source deep
learning software Caffe~\citep{JiaS14}.  We used OpenCL-Caffe with an
open source OpenCL back-end library called
ViennaCL~\citep{vcl}, which
provides efficient implementations of basic linear algebra operations
and some of the sparse matrix
functionality.

\subsection{Compressed Sparse Matrix}

In Section~\ref{sec:method}, we discussed how we could apply sparse
coding to fix the values of irrelevant learning weights of DNNs at
zero. However, if the zero values are explicitly stored in memory, the
model will use the same amount of memory as the original model without
compression. That is, for model compression, we need a special data
structure to store only the nonzero weight values in GPU~memory.

For our implementation, we have considered several popular formats,
in particular, DIA, ELL, CSR, and COO 
matrix formats to store the sparse
weight matrices~\citep{BelG08} (see Figure~\ref{fig:spformats} for
comparisons of the formats). Among these, DIA is suitable for the case
when nonzero values are at a small number of diagonals, and ELL is for
the case when matrix rows have similar numbers of nonzero
entries. Since there is usually no particular pattern of zero entries
in weight matrices produced by sparse coding, we have concluded that
these formats are unsuitable.

The compressed sparse row (CSR) format is probably the most popular
format for representing unstructured sparse matrices. This format
stores column indices and nonzero values in \texttt{indices} and
\texttt{data}, respectively, while in \texttt{ptr} it stores the
indices where new rows begin. Compared to DIA and ELL, the CSR format
can store variable numbers of nonzeros in rows efficiently. The
coordinate (COO) format is similar to CSR, but operations on COO can
be made simpler as it also stores row indices in \texttt{row} for
every nonzero entry. The extra storage required by COO for the row
indices appears to be less economical than CSR, as our target
platforms include small embedded systems. Therefore, we conclude that
the CSR format will be the best for representing compressed sparse
weight matrices in small, GPU-enabled devices. In ViennaCL, the CSR
format is implemented as the \texttt{C++} class
\texttt{compressed\_matrix}, and we have adapted this class to
implement our new matrix operations for forward/backward passes of
deep neural networks.

\begin{figure}[H]
Matrix:
$$
  A = \begin{bmatrix}
    1 & 7 & 0 & 0 \\
    0 & 2 & 8 & 0 \\
    5 & 0 & 3 & 9 \\
    0 & 6 & 0 & 4
\end{bmatrix}
$$

Compressed Matrix Representations:
\\~\\
(i) DIA format:
\begin{align*}
  \text{data} = \begin{bmatrix} * &  1 &  7 \\ * & 2 & 8 \\ 5 & 3 & 9 \\ 6 & 4 & * \end{bmatrix}, \;\;
  \text{offsets} =[-2\;\; 0\;\; 1]
\end{align*}
\\

(ii) ELL format:
\begin{align*}
 \text{data} = \begin{bmatrix}  1 &  7 & * \\ 2 & 8 & * \\ 5 & 3 & 9 \\ 6 & 4 & * \end{bmatrix}, \;\;
 \text{indices} = \begin{bmatrix}  0 &  1 & * \\ 1 & 2 & * \\ 0 & 2 & 3 \\ 1 & 3 & * \end{bmatrix}
\end{align*}
\\

(iii) CSR format:
\begin{align*}
  \text{row\_ptrs} &=     [\; 0\;\; 2\;\; 4\;\; 7\;\; 9\; ] \\
  \text{col\_indices} &= [\; 0\;\; 1\;\; 1\;\; 2\;\; 0\;\; 2\;\; 3\;\; 1\;\; 3\; ]\\
  \text{elements}    &= [\; 1\;\; 7\;\; 2\;\; 8\;\; 5\;\; 3\;\; 9\;\; 6\;\; 4\; ]
\end{align*}
\\

(iv) COO format:
\begin{align*}
  \text{row} &=     [\; 0\;\; 0\;\; 1\;\; 1\;\; 2\;\; 2\;\; 2\;\; 3\;\; 3\; ]\\
  \text{indices} &= [\; 0\;\; 1\;\; 1\;\; 2\;\; 0\;\; 2\;\; 3\;\; 1\;\; 3\; ]\\
  \text{data}    &= [\; 1\;\; 7\;\; 2\;\; 8\;\; 5\;\; 3\;\; 9\;\; 6\;\; 4\; ]
\end{align*}

\caption{Compressed matrix representation of a simple matrix
  $A$. The entries marked by $*$ are padding entries~(\citep{BelG09}, Figure~2).
\label{fig:spformats}}
\end{figure}

\subsection{Sparse Matrix Multiplication in OpenCL}

For training a deep neural network, we need efficient sparse matrix
operations to deal with the compressed sparse weight matrices in the
CSR format. To simplify notations, let us use $W$ to denote the sparse
weight matrix that coordinates the transfer between two consecutive
DNN layers, we call the bottom and the top layers (bottom
$\rightarrow$ top is the forward pass direction). Denoting by $X_B$
and $X_T$ the input from the bottom layer and the output to the top
layer respectively and following the shapes and orders of matrices
according to the original implementation of Caffe, we can summarize
the matrix multiplications needed for the forward and the backward
propagation steps as follows:\medskip
\begin{align*}
  \text{Forward} \;&:\; X_T \leftarrow X_B W' \;\; \text{(dense$\times$compressed$'$)}, \\ 
  \text{Backward} \;&:\; \frac{\partial Loss}{\partial X_B} \leftarrow \frac{\partial Loss}{\partial X_T} W \;\; \text{(dense$\times$compressed)}.
\medskip
\end{align*}

Here, $X_B$ and $\frac{\partial Loss}{\partial X_T}$ are typically
dense matrices, and therefore we essentially need two types of
operations: dense$\times$compressed$'$ ($D\times C'$ in short), where
$'$ is the matrix transpose operation, and dense$\times$compressed
($D\times C$ in short).  \textls[-15]{Unfortunately, these operations are not
available in the current version of ViennaCL (accessed on 19 October 2018): 
there exist $C\times D$ and $C\times D'$ operations in ViennaCL, so we
could use a workaround $(C\times D')' = D\times C'$, but this requires
extra memory space for transposing the result of $C\times
D'$. Furthermore, such a workaround is not available for $D\times C$ since
$(D\times C)' = C' \times D'$, and the transpose operation for
compressed sparse matrices ($C'$) is not available in ViennaCL. As a
solution, we provide two new dense-compressed matrix multiplications
accelerated by OpenCL}.

\subsubsection{Dense $\times$ Compressed$'$}

\begin{figure}[H]
\centering
\includegraphics[width=0.9\linewidth]{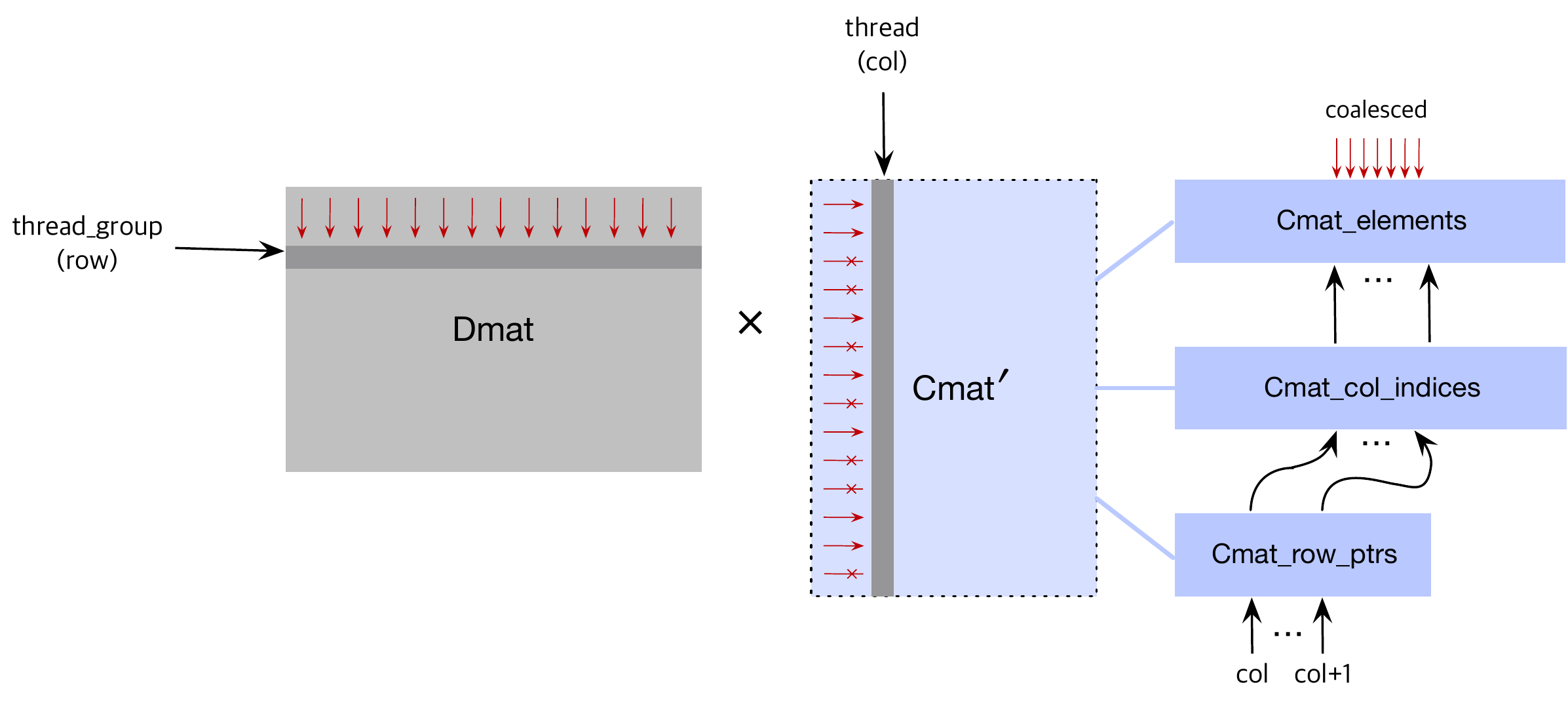}
{\begin{lstlisting}
// split work row to thread groups
for (unsigned int row = get_group_id(0); row < result_row_size; row += get_num_groups(0)) {
  // split result cols between threads in a thread group
  for (unsigned int col = get_local_id(0); col < result_col_size; col += get_local_size(0)) {
    double r = 0;
    unsigned int col_start = Cmat_row_ptrs[col]; // treat Cmat as it were transposed
    unsigned int col_end = Cmat_row_indices[col+1];
    for (unsigned int k = col_start; k < col_end; k++) {
      unsigned int j = Cmat_col_indices[k];
      double x = Dmat[row*Dmat_internal_cols + j]; 
      double y = Cmat_elements[k];
      r += x*y;
    }
    result[row*result_internal_cols + col] = r;
}}
\end{lstlisting}}

\caption{OpenCL kernel for the Dense(\texttt{Dmat}) $\times$ Compressed(\texttt{Cmat})$'$
  operation.\label{fig:DST} }
\end{figure}

In our implementation, this operation is used for computing forward
propagation. This operation is well suited for GPU acceleration, since
the compressed matrix stores nonzero elements rowwise, where we access
the nonzero elements stored in the compressed matrix in a rowwise
fashion to compute the inner product between a row of the dense matrix
and a column of the compressed$'$ matrix. Figure~\ref{fig:DST} shows a
sketch of the operation and our OpenCL kernel code to perform this
multiplication on GPUs: an OpenCL thread group is assigned to each row
of \texttt{Dmat} (dense matrix), and multiple threads in the group
will handle the assigned columns, one thread per column. For each
(row, col) pair, an inner product between the row in \texttt{Dmat} and
the column in \texttt{Cmat}$'$ (compressed$')$ will be computed. In
our discussion hereafter, we assume that \texttt{Dmat} and
\texttt{result} matrices are stored in a rowwise fashion. The column
memory access of \texttt{Cmat}$'$ equals to the row access of
\texttt{Cmat}, and therefore we can use \texttt{Cmat\_row\_ptrs} to
enumerate the nonzero elements corresponding to the variable
\texttt{col}. Each thread can access only the required nonzero
elements in consecutive memory locations, and therefore thread memory
access will be coalesced, leading to efficient parallelism on GPUs.

\subsubsection{Dense $\times$ Compressed}

\begin{figure}[H]
\centering
\includegraphics[width=.9\linewidth]{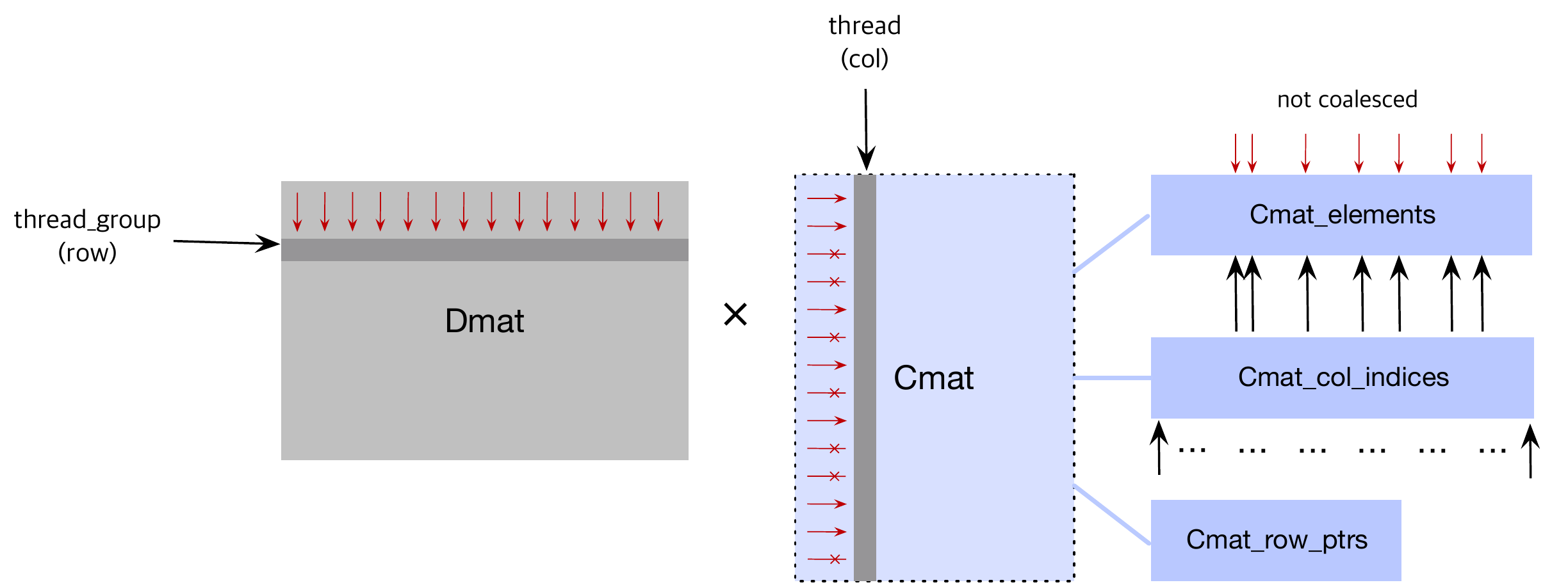}
\begin{lstlisting}
// split work rows to thread groups
for (unsigned int row = get_group_id(0); row < result_col_size; row += get_num_groups(0)) {
  // split result cols between threads in a thread group
  for (unsigned int col = get_local_id(0); col < result_row_size; col += get_local_size(0)) {
    double r = 0;
    unsigned int row_idx = 0;
    for (unsigned int k = 0; k < Cmat_nnz; k++) {
      if(row_idx < (Dmat_col_size-1) && k == Cmat_row_indices[row_idx+1]) row_idx++;
      if(col == Cmat_col_indices[k]) {
        double x = Dmat[ row*Dmat_internal_cols + row_idx];
        double y = Cmat_elements[k];
        r+=x*y;
        k = Cmat_row_indices[++row_idx]-1;
      }
    }
    result[row*result_internal_cols + col] = r;
}}
\end{lstlisting}
\caption{OpenCL kernel for the Dense(\texttt{Dmat}) $\times$ Compressed(\texttt{Cmat})
  operation.\label{fig:DS} }
\end{figure}

This dense-compressed matrix operation is required for backward
propagation. Unlike the previous one, this operation is not ideally
suited for GPU parallelism since the OpenCL kernel needs to access the
\texttt{Cmat} matrix columnwise, while the nonzero entries of the
compressed matrix are stored in a rowwise fashion. Unless we store an
extra array for nonzero entries stored columnwise in the compressed
matrix data structure, it is unavoidable that the memory access
pattern will not coalesce. Still, we can design an OpenCL kernel
to parallelize the computation for each (row, col) pair, as shown in
Figure~\ref{fig:DS}.

\subsection{Prox Operator in OpenCL}

The proximal operator discussed in Section~\ref{sec:proximal} can be
performed in parallel since the outcome can be computed elementwise,
as shown in Equation~\eqref{eq:prox}. ViennaCL has a collection of kernels for
matrix unary elementwise operations, which provides a basis for
implementing our own kernel. Our code in Figure~\ref{fig:prox} assigns
thread groups to rows and threads in groups to columns, similar to the
two OpenCL kernels we discussed above.

\begin{figure}[H]
\centering
\begin{lstlisting}
// split work rows to thread groups
for (unsigned int row = row_gid; row < result_row_size; row += get_num_groups(0)) {
  // split result cols between threads in a thread group
  for (unsigned int col = col_gid; col < result_col_size; col += get_local_size(0)){
     __global double *elem = &A[row*result_col_size + col];
     *elem = min(max(*elem-lambda*learning_rate,0.0),*elem+lambda*learning_rate);
  }
}
\end{lstlisting}
\caption{OpenCL kernel for the proximal operator, where the matrix
  \texttt{A} contains $w^k - \eta G(w^k)$.\label{fig:prox} }
\end{figure}

\section{Experiments}



To show the effectiveness of our compression method for deep neural
networks, we tested our method on four popular convolutional neural
networks on image recognition tasks: (i) Lenet-5 on the MNIST
dataset~\citep{LecB98} and (ii) AlexNet~\citep{KriS12}, (iii)
VGG16~\citep{SimZ15}, and (iv) ResNet-32~\citep{HeZ16} on the CIFAR-10
dataset~\citep{CIFAR10}. In both datasets, we have used the original
split of training and test data (MNIST has $28\times 28$ grey-scaled
images with no. of train/test examples = $60k/10k$, and CIFAR-10 has
\mbox{$32\times 32$ color images} with train/test = $50k/10k$). We fixed the
number of training updates to $60k$ with a minibatch size of $128$ (so
that the effective number of example iterations will be
$60k \times 128$) since the number was large enough for the resulting
neural networks to reached known top prediction accuracy. We compared
our proposed method based on $\ell_1$-regularized sparse coding
(denoted by SpC), to the existing pruning method~\citep{HanP15}
(denoted by Pru) based on thresholding and retraining and the
state-of-the-art method~\citep{CarI18} (denoted by MM) based 
on
the method of multipliers optimization algorithm. To initialize weight
values, we have used the initialization from \citet{HeZ15}, which is
known to work well with neural networks containing the ReLU
activation.

Our implementation is available as open-source
software (\url{https://github.com/sanglee/caffe-mc-opencl}).

\subsection{Comparison of Training Algorithms}

Our algorithms Prox-RMSProp (Algorithm~\ref{alg:prox-rmsprop}) and
Prox-ADAM (Algorithm~\ref{alg:prox-adam}) use random initialization
and random minibatch examples in weight updates. Furthermore, model
compression will impose statistical bias compared to the reference
models without compression. Hence, it is natural to expect some
variation if we repeat training with different random seeds.

\begin{figure}[H]
  \centering
  \subfloat[Prox-RMSProp]{\includegraphics[width=.5\linewidth]{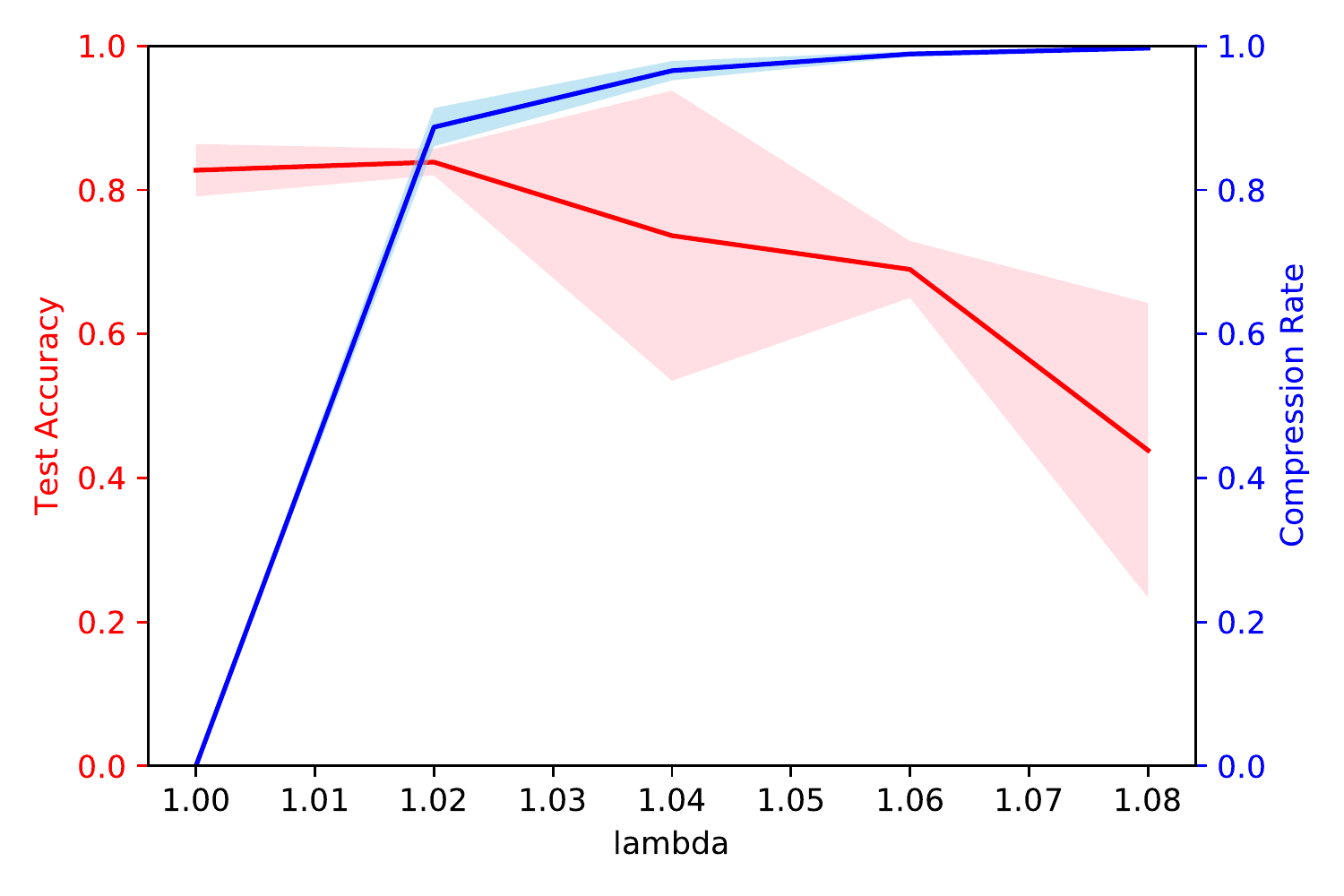}\label{fig:scomp-a}}
  ~
  \subfloat[Prox-ADAM]{\includegraphics[width=.5\linewidth]{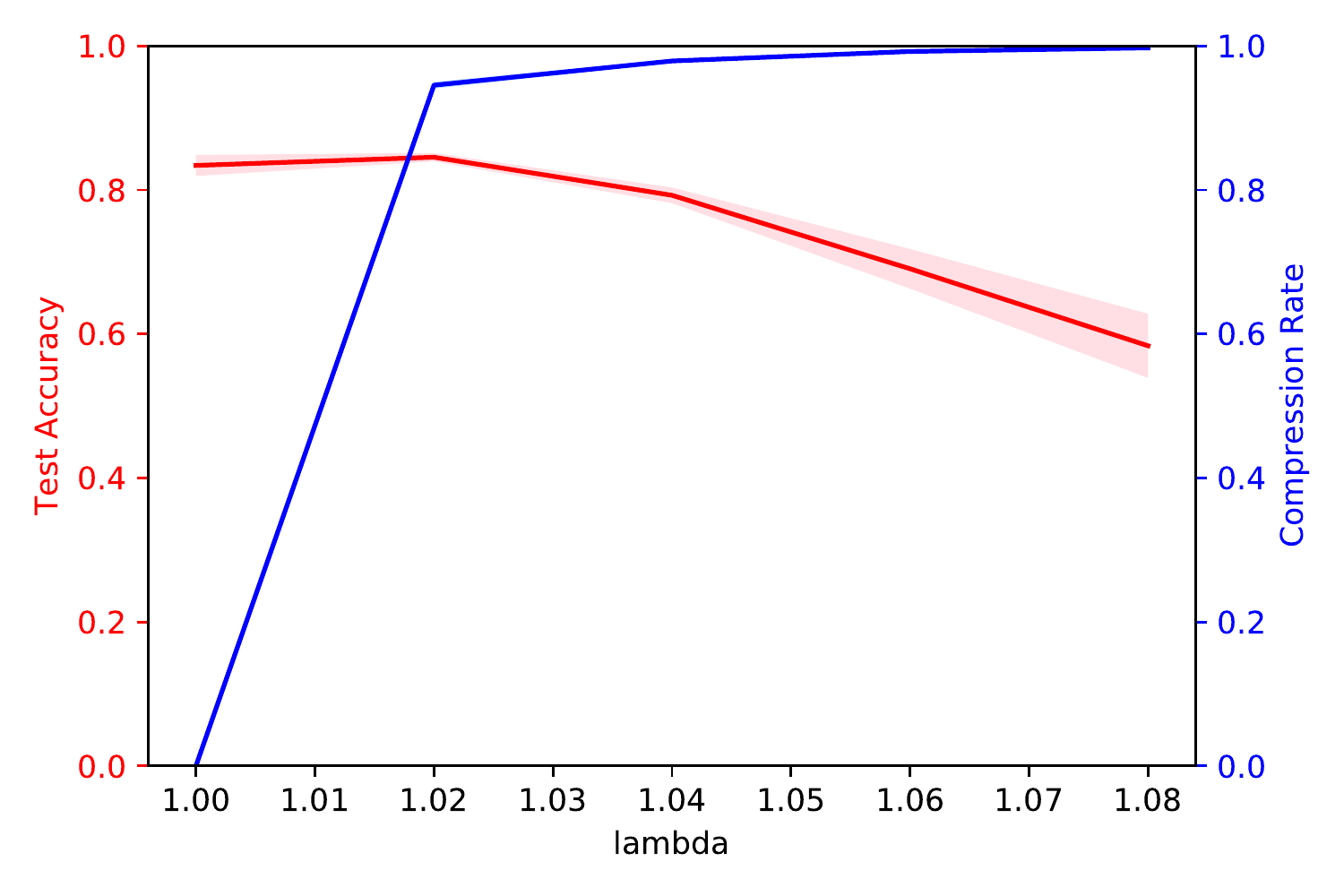}\label{fig:scomp-b}}

  \caption{Comparison of our suggested algorithms (\textbf{a}) Prox-RMSProp and
    (\textbf{b}) Prox-ADAM, on VGGNet training with the CIFAR-10
    dataset. Prox-ADAM produced more stable trained models in terms of
    test accuracy and compression rate.}
  \label{fig:solver-comp}
\end{figure}

Here, we compare our two algorithms in order to investigate how much
variation they will exhibit in training, regarding test accuracy and
compression rate (the ratio of the number of zero entries to the total
number of learning parameters). We have trained Lenet-5 (MNIST),
AlexNet (CIFAR-10), VGGNet (CIFAR-10), ResNet-32 (CIFAR-10) multiple
times with different random seeds. Amongst the experiments, our
algorithms showed the most distinctive characteristics on VGGNet, as
shown in Figure~\ref{fig:solver-comp}. Overall, Prox-ADAM showed
smaller variance in both test accuracy and compression rate than
Prox-RMSProp. This behavior is expected since the search directions
produced by Prox-ADAM is more stable than those of Prox-RMSProp: in
the former, the direction is a composition of a momentum direction
(which brings stability) and a minibatch gradient direction, whereas
in the latter it is solely based on a noisy minibatch gradient. For
this reason, we chose Prox-ADAM for the rest of our~experiments.

\subsection{Compression Rate and Prediction Accuracy}

In the training problem with sparse coding~\eqref{eq:min}, we use the
$\ell_1$-regularizer $\Psi(w) = \lambda \|w\|_1$ where $\lambda>0$ is
a hyperparameter that determines the amount of compression (higher
$\lambda$ gives larger numbers of zero entries in $w$ and thereby
higher compression rate).

Figure~\ref{fig:crate} shows test accuracy and compression rate of the
four network-dataset pairs with respect to the change of the $\lambda$
parameter value, for our sparse coding approach
(Figure~\ref{fig:crate}a, SpC) and the existing pruning approach
(Figure~\ref{fig:crate}b, Pru). The test accuracy values of the
reference networks (trained without sparse coding) are shown as the
horizontal dotted lines. \textbf{SpC}: at small $\lambda$ values, we
can see that the test accuracy of compressed models is usually better
than that of the reference models. This may happen when the reference
model is overfitting the data and $\ell_1$ regularization is
mitigating the effect. When the test accuracy values of the compressed
models were similar to those of the reference models (crossings of
vertical and horizontal dotted lines), our method was able to remove
about $90\%$ of the weights (one exception was ResNet-32, for which
both Pru and SpC without retraining showed poor compression). The plots
also indicate that further compression will be possible if we are
ready to sacrifice prediction accuracy by a small
margin. \textbf{Pru}: test accuracy values drop much more rapidly as
we increase the compression rate, compared to our sparse coding
approach. Similar test accuracy to the reference model has been
achieved with about $40\%$ of compression in case of Lenet-5, but no
compression was possible for AlexNet, VGGNet, and ResNet-32 if we
wanted to achieve similar test accuracy to the reference model.
\begin{figure}[H]
  \centering
  \subfloat[Accuracy vs. Compression Rate (Our Approach: SpC)\label{fig:crate-a}]{%
    \includegraphics[width=.95\textwidth]{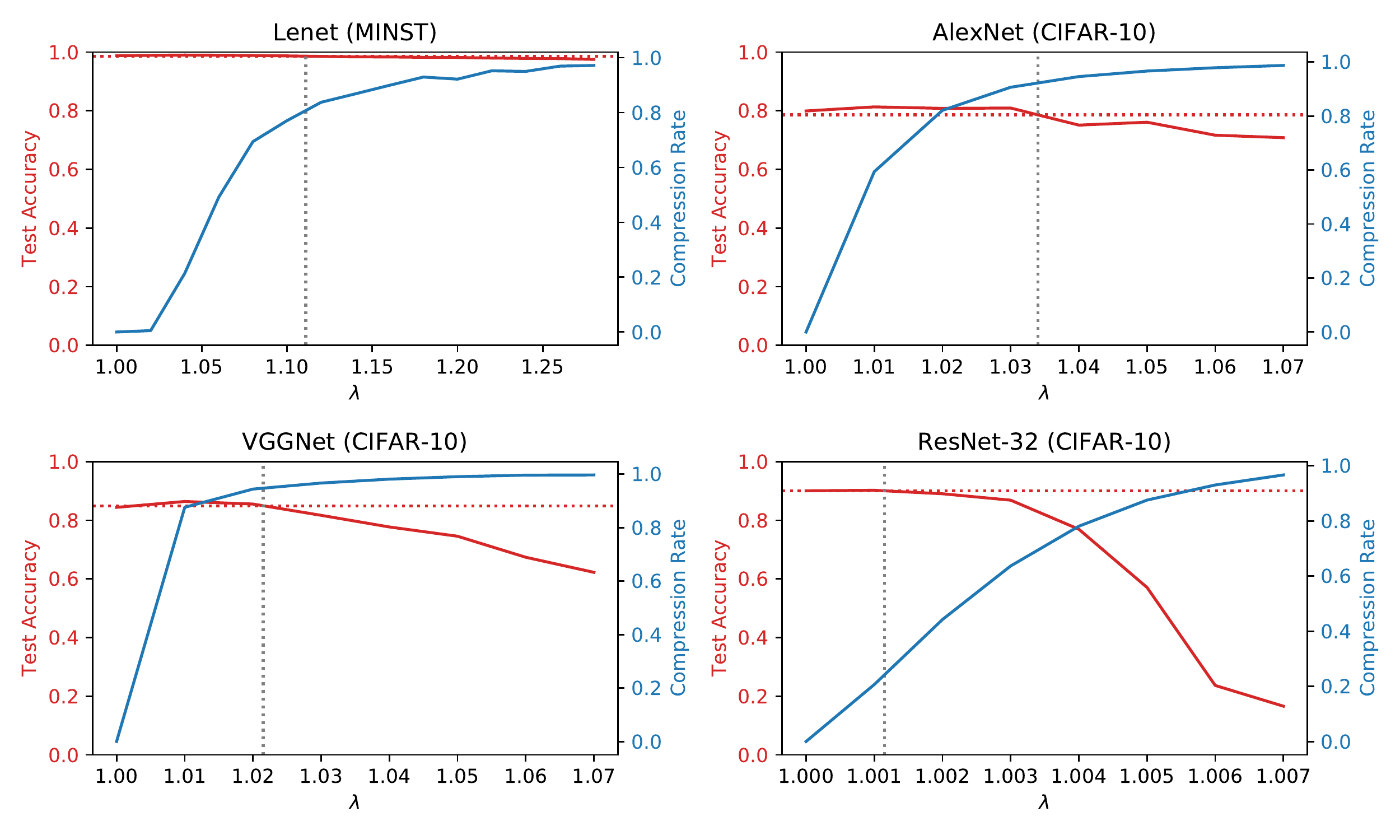}}

  \caption{\textit{Cont}.}
  \label{fig:crate}
\end{figure}
\begin{figure}[H]\ContinuedFloat
  \centering
  \subfloat[Accuracy vs. Compression Rate (Pruning Approach: Pru)  \label{fig:crate-b}]{%
    \includegraphics[width=.95\textwidth]{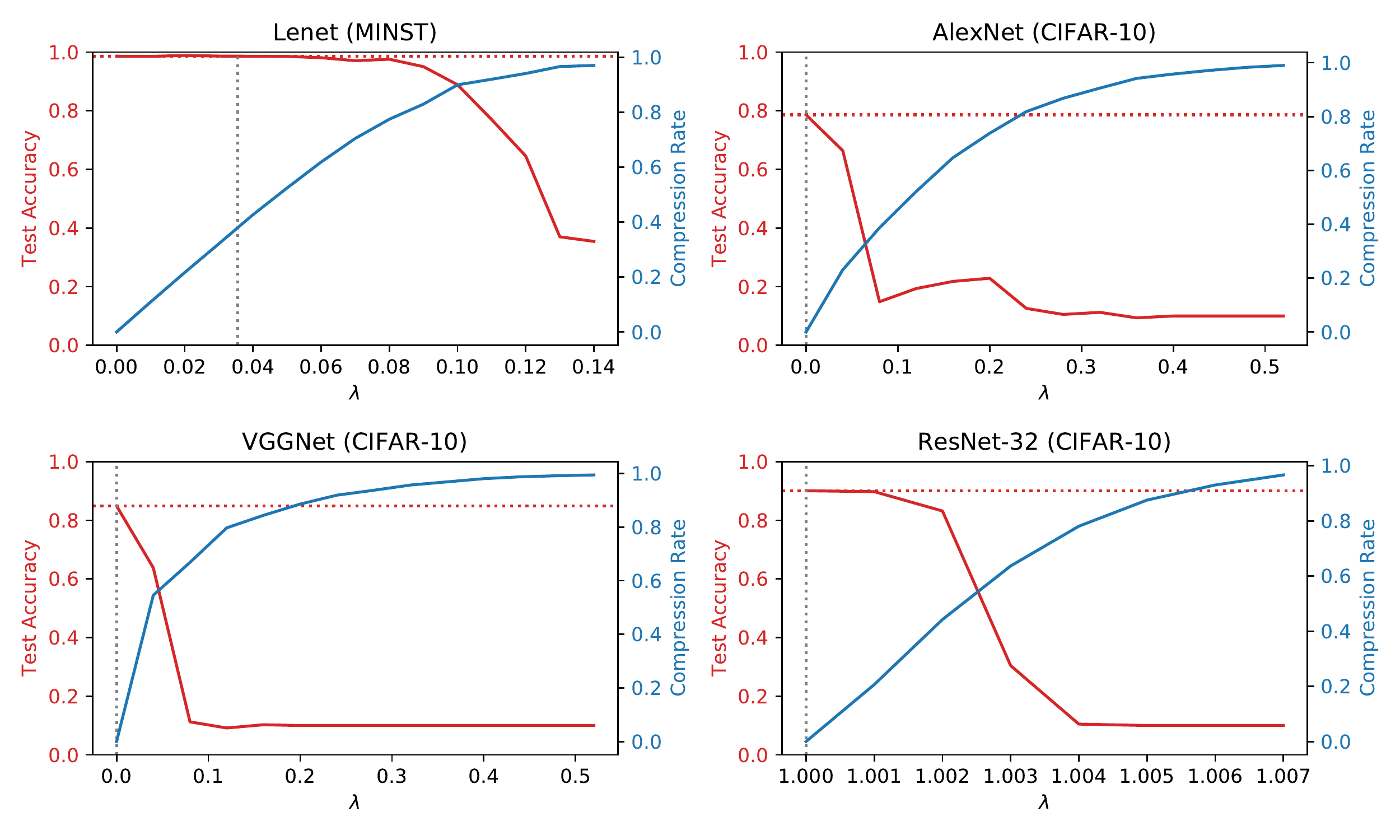}}

  \caption{Compression rate and test accuracy with respect to the
    change of the $\lambda$ parameter value, for (\textbf{a}) our sparse coding
    approach Spc and (\textbf{b}) the existing pruning approach. Higher
    $\lambda$ values result in more compression with a decrease in
    prediction accuracy. (\textbf{a}) Spc: using sparse coding, about $90\%$ of
    model parameters can be compressed achieving similar prediction
    accuracy to the reference models (test accuracy is shown as the
    horizontal dotted lines, and their crossings with accuracy curves
    of compressed models are marked as vertical dotted lines); (\textbf{b})
    Pru: the existing pruning approach does not compress the networks
    as much: about $40\%$ compression for Lenet-5 but no compression
    for the other networks at the similar accuracy level to the
    reference model.}
  \label{fig:crate}
\end{figure}
\subsection{The Effect of Retraining}

Compressed models can be debiased by means of an extra training the
compressed networks, where the weights at the zero value are fixed and
not updated during retraining.  Figure~\ref{fig:spacc} shows the
prediction accuracy of the neural networks at different compression
rates, comparing models produced by SpC and Pru, and their retrained
versions, SpC(Retrain) and Pru(Retrain).

We can see that retraining is indeed a necessary step for the pruning
approach (Pru) as previously known~\citep{HanP15}, since otherwise
compressed models show impractically low prediction performance. The
pruning method with retraining, Pru(Retrain), shows similar prediction
accuracy to our sparse coding method SpC (without retraining) up to
a moderate compression rate. However, when the compression rate is very
high, our method SpC clearly outperforms Pru(Retrain). In addition,
retraining improves the prediction accuracy of our method SpC even
further, especially when the compression rate is very high, which can be
seen from the solid and dotted vertical lines where SpC and
SpC(Retrain), respectively, achieve at least $99\%$ of the reference
accuracy with maximal compression. Therefore, for our sparse coding
method SpC, retraining can be used when we aim for very high
compression.

Table~\ref{tab:summary} summarizes the compression and prediction
results of all cases.
\begin{figure}[H]
  \centering
  \includegraphics[width=.9\linewidth]{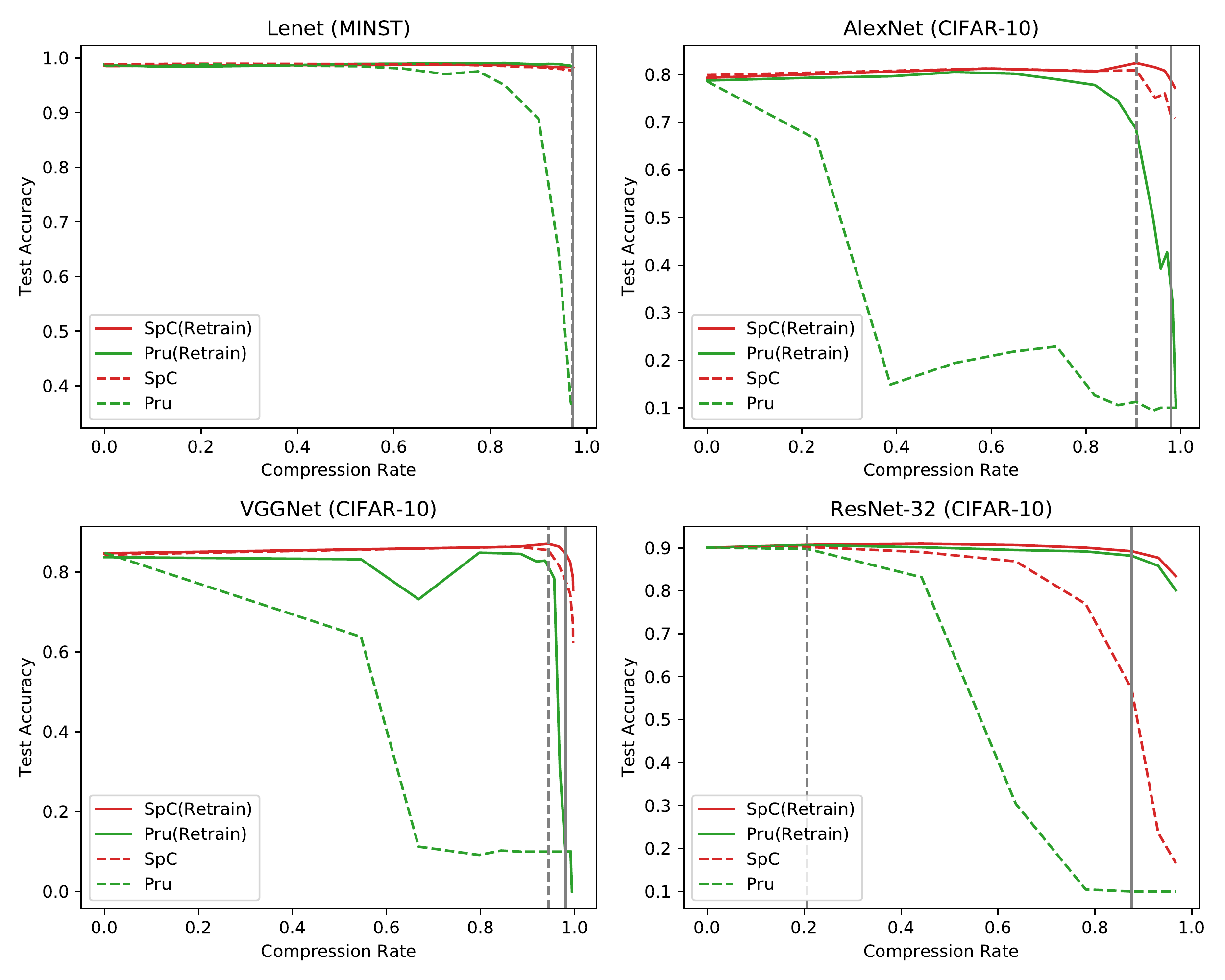}
  \caption{The effect of retraining. The maximal compression rates
    where our methods SpC 
    and SpC(Retrain) achieve at least $99\%$ of
    the reference prediction accuracy are indicated by vertical lines
    (solid: SpC, dotted: SpC(Retrain)). Retraining is indeed required
    for Pru since otherwise high compression rate is not possible to
    achieve good prediction performance. On the contrary, our method
    SpC without retraining can achieve high accuracy at high
    compression. Retraining for SpC can be considered when we aim for
    very high compression. }
  \label{fig:spacc}
\end{figure}
\unskip

\begin{table}[H]
\centering
\caption{Summary of network compression results. (Pru: network pruning, SpC: our method)}
\label{tab:summary}
\centering
\begin{tabular}{cccccc}
\toprule
\multicolumn{2}{c}{\textbf{Network}} &  \textbf{Lenet-5} & \textbf{AlexNet} & \textbf{VGGNet} & \textbf{ResNet-32}\\
\midrule
  \multicolumn{2}{c}{Data} &  MNIST & CIFAR-10 & CIFAR-10 & CIFAR-10 \\
  \multicolumn{2}{c}{Ref. Accuracy} & $98.61\%$ & $78.61\%$ & $84.88\%$ & $90.05\%$ \\
  \midrule
\multirow{2}{*}{Pru} & Accuracy & $37.01\%$ & $11.24\%$ & $10.00\%$ & $89.76\%$ \\
& Compression Rate   & $0.97~(29\times)$ & $0.91~(10\times)$ & $0.94~(15\times)$ & $0.21~(1\times)$ \\
\multirow{2}{*}{Pru (Retrain)} & Accuracy   & $98.19\%$ & $42.67\%$ & $10.00\%$ & $88.16\%$ \\
& Compression Rate   & $0.97~(33\times)$ & $0.97~(35\times)$ & $0.98~(51\times)$ & $0.88~(8\times)$ \\\midrule              
\multirow{2}{*}{SpC} & Accuracy & $97.78\%$ & $80.93\%$ & $85.53\%$ & $90.22\%$ \\
                     & Compression Rate   & $0.97~(32\times)$ & $0.91~(10\times)$ & $0.94~(17\times)$ & $0.21 (1\times)$ \\
\multirow{2}{*}{SpC (Retrain)} & Accuracy   & $98.29\%$ & $78.84\%$ & $84.63\%$ & $89.22\%$ \\
    & Compression Rate   & $0.97 (35\times)$ & $0.98~(47\times)$ & $0.98~(53\times)$ & $0.88~(8\times)$ \\ \bottomrule
\end{tabular}
\end{table}

\subsection{Comparison to the State-of-the-Art Approach}

We also compare our method to the state-of-the-art network pruning
approach based on the method of multipliers~\citep{CarI18}, which we
refer to as MM. In this method, one first duplicates the learning
parameters $w$ with an auxiliary variable $\theta$, to convert the
training regularization problem~\eqref{eq:min} into the following
constrained optimization problem:
\begin{equation}\label{eq:mm}
 \min_{w\in\R^p, \theta\in\R^p} \;\; L(w) + \alpha \Psi(\theta), \;\; \text{subject to} \;\; w =  \theta, \enspace
\end{equation}
where $L(w) = \frac{1}{n} \sum_{i=1}^n \ell(w;x_i,y_i)$ is the total
loss and $\Psi(\theta)$ is a regularizer. Then MM constructs an
augmented Lagrangian function
\begin{equation}\label{eq:mm.2}
  \mathcal L(w,\theta,\lambda;\mu) = L(w) + \frac{\mu}{2} \|w-\theta\|^2 - \lambda^T (w-\theta) + \Psi(\theta). \enspace
\end{equation}
Here, $\lambda \in \R^p$ is the Lagrange multipliers and $\mu > 0$ is
an auxiliary parameter. Then, MM iterates (i) the minimization of
$\mathcal L$ (in terms of $w$ and $\theta$ in an alternating fashion)
and (ii) a gradient ascent step of $\mathcal L$ in $\lambda$, while
driving $\mu \to \infty$.

Our approach has several benefits over the MM method. First, MM
requires a pre-trained model to start network pruning, whereas our
method can start from random weights. Second, due to the duplication
of learning weights $w\in\R^p$ by $\theta$ in Equation~\eqref{eq:mm} and the
introduction of the Lagrange multipliers $\lambda \in \R^p$, MM
requires about double amount of memory for training than our method
(ours: $(w,\nabla_w L(w))$ and MM:
$(w,\nabla_w L(w), \theta, \lambda)$). Third, the convergence of MM is
quite sensitive to how we control learning rates and the auxiliary
parameter $\mu \to \infty$ in the augmented Lagrangian, where our
method has no such sensitivity.

Table~\ref{tab:mm-prox} shows the comparison between our method SpC
and the state-of-the-art method MM, regarding two convolutional neural
networks tested in the original paper~\citep{CarI18}. Note that the
comparison would be more favorable to MM because MM is allowed (i) to
start from the state-of-the-art pretrained models and (ii) to use
different solvers and auxiliary parameter control strategies as in the
original paper~\citep{CarI18}. Nevertheless, our method SpC shows a
competent compression performance even though its compressed learning
starts from random weight values.

\begin{table}[H]
\centering
\caption{Comparison between our method SpC and the
state-of-the-art MM. For MM, we used the hyperparameters
and the best pretrained models as a starting point
following the original paper~\citep{CarI18}. Note that our
method SpC starts from random weights and does not require
complicate control of the auxiliary
parameter $\mu$.}
\label{tab:mm-prox}
\footnotesize
\centering
\scalebox{.95}[0.95]{\begin{tabular}{ccccc}
\toprule
  \textbf{Network (Data)} & \multicolumn{2}{c}{\textbf{Lenet-5 (MNIST)}} & \multicolumn{2}{c}{\textbf{ResNet32 (CIFAR10)}}\\\midrule
  Method       & SpC & MM & SpC & MM  \\
  \midrule
  Pretrained Model & - & Required (Test Acc = 99.1) & - & Required (Test Acc = 92.28)\\
  Solver & Prox-Adam & SGD with Momentum &  Prox-Adam & Nesterov \\
  Aux. Parameter ($\mu$) & - & $9.76\times 10^{-5}$ ($\times 1.1$ per 4k iter) & - & $10^{-3}$($\times 1.1$ per 2k iter)\\
  \midrule
  Accuracy & 97.25\% & 97.65\% & 89.22\% & 92.37\%\\
  Compression Rate & 0.98 & 0.98 & 0.88 & 0.85 \\
\bottomrule
\end{tabular}}
\end{table}

Figure~\ref{fig:spacc} shows the convergence behavior of SpC and
MM. The compression is performed every update in SpC and every 4k
updates in MM, and therefore the compression rate convergence may look
smoother in MM. (In fact, it is another tuning parameter in MM how
often it performs weight compression, and we found that the algorithm
is quite sensitive to different choices.) However, our method SpC
achieves the top compression rate and test accuracy much faster than
MM. Therefore, our method suits better for embedded systems where we
can afford only a few thousand training iterations due to time and
energy constraints.

\begin{figure}[t!]
	\centering
	\includegraphics[width=.9\linewidth]{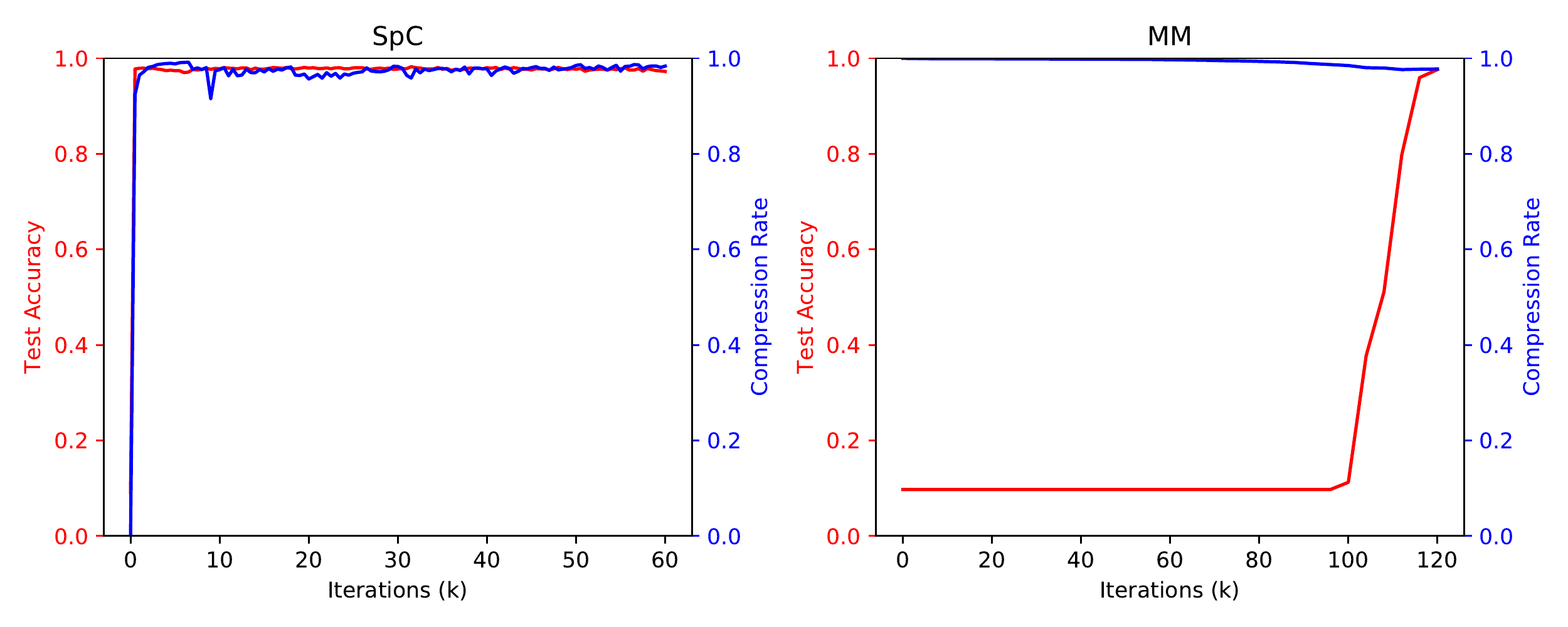}
	\caption{Convergence behavior of the our method SpC (left) and the state-of-the-art MM (right) in
          terms of compression rate and test accuracy on Lenet-5
          (MNIST).  Our method SpC reaches top accuracy and
          compression much faster than MM. Note that the last
          iterations are 60k (SpC) and 120k (MM).  }
	\label{fig:spacc}
\end{figure}

\subsection{Performance on Embedded Systems}

We also tested our suggested framework in an embedded system, which
has a similar spec to the Samsung Galaxy S7 smartphone (Seoul, Korea). Our test
system has six-core 64-bit ARM CPU, 4 GB main memory, ARM Mali-T860 GPU,
OpenCL 1.2 support, and Ubuntu 16.04 operating system. In this case,
we are interested in the inference time comparison between full
reference models and compressed models, since in the latter the
weights are stored in a compact form and will be computed with sparse
matrix operations, and therefore some speedups can be expected.

Table~\ref{tab:embedded} shows the results on the test embedded
system, also making a comparison to the results on a workstation with
an NVIDIA GPU. Despite that we have achieved some speedup, it seems to
fall short considering the size of compressed models. One reason is
that the compressed convolution filters have irregular nonzero
patterns for which full GPU acceleration is difficult. We plan to
investigate this issue further in our future research.

\begin{table}[H]
	\begin{center}
		\caption{Inference speedups by model compression (Lenet-5, MNIST). }
		\label{tab:embedded}
		\centering
		\begin{tabular}{ccccc}\toprule
			\textbf{GPU} & \multicolumn{2}{c}{\textbf{NVIDIA GTX 1080 TI}} & \multicolumn{2}{c}{ \textbf{ARM Mali-T860} }\\\midrule
                        Compression  & Yes & No  & Yes  & No \\ 
                        Model Size & 148KB & 5.0MB & 148KB & 5.0MB \\\midrule
			Inference Time & 8572~ms  & 16,977~ms & 506,067~ms  & 606,699~ms \\ 
			Speed up & \multicolumn{2}{c}{2$\times$} &\multicolumn{2}{c}{ 1.2$\times$} \\\bottomrule
		\end{tabular}
\end{center}
\end{table}

\section{Conclusions}

In this paper, we proposed an efficient model compression framework
for deep neural networks based on sparse coding with $\ell_1$
regularization, proximal algorithms, debiasing, and compressed
computation of sparse weights using OpenCL. We believe that our method
will be more versatile for embedded system applications than the
previous methods as it does not require a pre-trained model as an
input, while it can produce competent compressed models.

\vspace{6pt} 



 \authorcontributions{Conceptualization, S.L.; formal analysis, S.L. and J.L.; software, S.L and J.L.; writing -- original draft preparation, S.L.; writing--review and editing, S.L.}

\funding{This work was supported by the research fund of Hanyang University (HY-2017-N).}


\conflictsofinterest{The authors declare no conflict of interest.}



\clearpage 
\appendixtitles{yes} 
\appendix
\section{Layer-Wise Compression Rate}
Here, we check the compression rates of each layer of the three
networks, Lenet-5, AlexNet, and VGGNet when we apply our sparse coding
methods SpC and SpC(Retrain). We chose the regularization parameter
value $\lambda$ for each method so that the compressed models achieved
at least $99\%$ of the test accuracy of the reference model while
maximizing compression rates among trials.

Tables \ref{tab:lenet}--\ref{tab:resnet} show the results. We can again confirm our discussion
in the previous experiments that our method without retraining, SpC,
provides superb compression rate being on par with the reference model
in prediction accuracy. SpC showed compression rates: Lenet-5
($96.9\%$), AlexNet ($90.65\%$), VGGNet ($94.41\%$), and ResNet
($20.67\%$). Further compression was also possible with
retraining. SpC(Retrain) showed compression rates: Lenet-5
($97.16\%$), AlexNet ($97.88\%$), VGGNet ($98.13\%$), and ResNet
($87.50\%$), while prediction accuracy values of compressed models
were similar or slightly improved from those of SpC. From the
tables, we can also observe that the layers that are near the input
and the output layers are compressed less than the other layers in the
middle. One could use such information to redesign and optimize the
architecture of the neural networks for better computational
efficiency.


\begin{table}[H]
  \small
\caption{Layer-wise compression of lenet-5 (MNIST dataset). NNZ stands for number of non-zeros.}
\label{tab:lenet}
\centering
\begin{tabular}{ccccc}\toprule
\textbf{ Methods} & \multicolumn{2}{c}{ SpC } & \multicolumn{2}{c}{ \textbf{SpC(Retrain)} }\\\midrule
Layers & NNZ/Total Weights & Compression Rate & NNZ/Total Weights &
Compression Rate \\ \midrule
conv1 & 158/500  & 68.40\% (3$\times$) &  142/500  & 71.60\% (3$\times$) \\
conv2 & 2101/25,000  & 91.60\% (11$\times$) &  1750/25,000  & 93.00\% (14$\times$) \\
fc1 & 10,804/400,000  & 97.30\% (37$\times$) & 10,045/400,000  & 97.49\% (39$\times$) \\
fc2 & 270/5000  & 94.60\% (18$\times$) &  280/5000  & 94.40\% (17$\times$)  \\
\midrule
Total &  13,333/430,500 & 96.90\% (32$\times$) &  12,217/430,500 & 97.16\% (35$\times$) \\\midrule
Test Accuracy & 0.9778 @ $\lambda$ = 1.26 & Ref: 0.9861  & 0.9829 @ $\lambda$ = 1.28 & Ref: 0.9861 \\
\bottomrule
\end{tabular}
\end{table}
\unskip
\begin{table}[H]
 \small
\caption{Layer-wise compression of AlexNet (CIFAR-10 dataset).}
\label{tab:alexnet}
\centering
\begin{tabular}{ccccc}
\toprule
 \textbf{Methods} & \multicolumn{2}{c}{ \textbf{SpC} } & \multicolumn{2}{c}{ \textbf{SpC(Retrain}) }\\
 \midrule
  Layers & NNZ / Total Weights & Compression Rate & NNZ / Total Weights & Compression Rate \\ 
  \midrule
conv1 & 3922/7200  & 45.53\% (1$\times$) & 2054/7200  & 71.47\% (3$\times$) \\
conv2 & 76,321/307,200  & 75.16\% (4$\times$) & 19,464/307,200  & 93.66\% (15$\times$)\\
conv3 & 153,921/884,736  & 82.60\% (5$\times$) & 27,757/88,4736  & 96.86\% (31$\times$) \\
conv4 & 153,000/663,552  & 76.94\% (4$\times$) &  21,485/663,552  & 96.76\% (30$\times$) \\
conv5 & 516,47/442,368  & 88.32\% (8$\times$) &  15,690/442,368  & 96.45\% (28$\times$) \\
fc1 & 179,344/4,194,304  & 95.72\% (23$\times$) & 52,429/4,194,304  & 98.75\% (79$\times$) \\
fc2 & 84,495/1,048,576  & 91.94\% (12$\times$) & 18,841/1,048,576  & 98.20\% (55$\times$) \\
fc3 & 3701/10,240  & 63.86\% (2$\times$) & 2329/10,240  & 77.26\% (4$\times$)  \\
\midrule
Total &  706,351/7,558,176 & 90.65\% (10$\times$)  &  160,049/7,558,176 & 97.88\% (47$\times$)  \\
\midrule
Test Accuracy & 0.8093 @ $\lambda$ = 1.03 & Ref: 0.7861  & 0.7884 @ $\lambda$ = 1.06 & Ref: 0.7861 \\
\bottomrule
\end{tabular}
\end{table}
\unskip
\begin{table}[H]
\tiny
\caption{Layer-wise compression of VGGNet (CIFAR-10 dataset).}
\label{tab:vggnet}
\centering
\begin{tabular}{ccccc}\toprule
 \textbf{Methods} & \multicolumn{2}{c}{ \textbf{SpC} } & \multicolumn{2}{c}{ \textbf{SpC(Retrain)} }\\\midrule
 \textbf{Layers }& \textbf{NNZ/Total Weights} & \textbf{Compression Rate} & \textbf{NNZ/Total Weights} & \textbf{Compression Rate} \\ \midrule
conv1-1 & 1160/1728  & 32.87\% (1$\times$) & 757/1728  & 56.19\% (2$\times$) \\
conv1-2 & 18,904/36,864  & 48.72\% (1$\times$) & 9389/36,864  & 74.53\% (3$\times$) \\
conv2-1 & 47,497/73,728  & 35.58\% (1$\times$)  & 20943/73,728  & 71.59\% (3$\times$) \\
conv2-2 & 87,314/147,456  & 40.79\% (1$\times$) & 30,040/147,456  & 79.63\% (4$\times$) \\
conv3-1 & 133,402/294,912  & 54.77\% (2$\times$) & 40,039/294,912  & 86.42\% (7$\times$) \\
\bottomrule
\end{tabular}
\end{table}
\unskip
\begin{table}[H]\ContinuedFloat
\tiny
\caption{\textit{Cont}.}
\label{tab:vggnet}
\centering
\begin{tabular}{ccccc}\toprule
 \textbf{Methods} & \multicolumn{2}{c}{ \textbf{SpC} } & \multicolumn{2}{c}{ \textbf{SpC(Retrain)} }\\\midrule
 \textbf{Layers }& \textbf{NNZ/Total Weights} & \textbf{Compression Rate} & \textbf{NNZ/Total Weights} & \textbf{Compression Rate} \\ 
 \midrule
conv3-2 & 120,094/589,824  & 79.64\% (4$\times$)  & 30,997/589,824  & 94.74\% (19$\times$)  \\
conv3-3 & 94,612/589,824  & 83.96\% (6$\times$)  & 16,071/589,824  & 97.28\% (36$\times$) \\
conv4-1 & 164,660/1,179,648  & 86.04\% (7$\times$)  & 20,322/1,179,648  & 98.28\% (58$\times$) \\
conv4-2 & 133,944/2,359,296  & 94.32\% (17$\times$)   & 22,145/2,359,296  & 99.06\% (106$\times$) \\
conv4-3 & 59,355/2,359,296  & 97.48\% (39$\times$) & 28,173/2,359,296  & 98.81\% (83$\times$) \\
conv5-1 & 16,749/2,359,296  & 99.29\% (140$\times$)  & 21,349/2,359,296  & 99.10\% (110$\times$) \\
conv5-2 & 10,769/2,359,296  & 99.54\% (219$\times$) & 30,008/2,359,296  & 98.73\% (78$\times$) \\
conv5-3 & 10,987/2,359,296  & 99.53\% (214$\times$)  & 24,027/2,359,296  & 98.98\% (98$\times$) \\
fc1 & 4176/524,288  & 99.20\% (125$\times$) & 6072/524,288  & 98.84\% (86$\times$) \\
fc2 & 5915/1,048,576  & 99.44\% (177$\times$)  & 4007/1,048,576  & 99.62\% (261$\times$) \\
fc3 & 508/10,240  & 95.04\% (20$\times$)  & 223/10,240  & 97.82\% (45$\times$)  \\
\midrule
Total &  910,046/16,293,568 & 94.41\% (17$\times$)  &  304,562/16,293,568 & 98.13\% (53$\times$)  \\\midrule
Test Accuracy & 0.8553 @ $\lambda$ = 1.02 & Ref: 0.8488  & 0.8463 @ $\lambda$ = 1.04 & Ref: 0.8488 \\
\bottomrule
\end{tabular}
\end{table}
\unskip
\begin{table}[H]
\footnotesize
		\caption{Layer-wise compression of ResNet32 (CIFAR-10 dataset).}
		\label{tab:resnet}
		\centering
\scalebox{.98}[0.98]{\begin{tabular}{ccccc}
\toprule
	 \textbf{Methods} & \multicolumn{2}{c}{ \textbf{SpC} } & \multicolumn{2}{c}{ \textbf{SpC(Retrain)} }\\
	 \midrule
\textbf{ Layers} & \textbf{NNZ / Total Weights }& \textbf{Compression Rate }& \textbf{NNZ / Total Weights} & \textbf{Compression Rate} \\ \midrule
	conv1  &  379/432  & 12.27\% (1$\times$) &  139/432  & 67.82\% (3$\times$) \\
	conv1-1-1  &  1844/2304  & 19.97\% (1$\times$) &  327/2304  & 85.81\% (7$\times$) \\
	conv1-1-2  &  1870/2304  & 18.84\% (1$\times$) &  337/2304  & 85.37\% (6$\times$) \\
	conv1-2-1  &  1874/2304  & 18.66\% (1$\times$) &  334/2304  & 85.50\% (6$\times$) \\
	conv1-2-2  &  1873/2304  & 18.71\% (1$\times$) &  341/2304  & 85.20\% (6$\times$) \\
	conv1-3-1  &  1847/2304  & 19.84\% (1$\times$) &  330/2304  & 85.68\% (6$\times$) \\
	conv1-3-2  &  1872/2304  & 18.75\% (1$\times$) &  322/2304  & 86.02\% (7$\times$) \\
	conv1-4-1  &  1874/2304  & 18.66\% (1$\times$) &  363/2304  & 84.24\% (6$\times$) \\
	conv1-4-2  &  1852/2304  & 19.62\% (1$\times$) &  344/2304  & 85.07\% (6$\times$) \\
	conv1-5-1  &  1859/2304  & 19.31\% (1$\times$) &  355/2304  & 84.59\% (6$\times$) \\
	conv1-5-2  &  1835/2304  & 20.36\% (1$\times$) &  326/2304  & 85.85\% (7$\times$) \\
	conv2-1-1  &  3700/4608  & 19.70\% (1$\times$) &  666/4608  & 85.55\% (6$\times$) \\
	conv2-1-2  &  7316/9216  & 20.62\% (1$\times$) &  1108/9216  & 87.98\% (8$\times$) \\
	conv2-1-proj  &  467/512  & 8.79\% (1$\times$) &  225/512  & 56.05\% (2$\times$) \\
	conv2-2-1  &  7292/9216  & 20.88\% (1$\times$) &  1191/9216  & 87.08\% (7$\times$) \\
	conv2-2-2  &  7325/9216  & 20.52\% (1$\times$) &  1160/9216  & 87.41\% (7$\times$) \\
	conv2-3-1  &  7394/9216  & 19.77\% (1$\times$) &  1198/9216  & 87.00\% (7$\times$) \\
	conv2-3-2  &  7371/9216  & 20.02\% (1$\times$) &  1160/9216  & 87.41\% (7$\times$) \\
	conv2-4-1  &  7323/9216  & 20.54\% (1$\times$) &  1222/9216  & 86.74\% (7$\times$) \\
	conv2-4-2  &  7368/9216  & 20.05\% (1$\times$) &  1200/9216  & 86.98\% (7$\times$) \\
	conv2-5-1  &  7265/9216  & 21.17\% (1$\times$) &  1223/9216  & 86.73\% (7$\times$) \\
	conv2-5-2  &  7303/9216  & 20.76\% (1$\times$) &  1179/9216  & 87.21\% (7$\times$) \\
	conv3-1-1  &  14,757/18,432  & 19.94\% (1$\times$) &  2411/18,432  & 86.92\% (7$\times$) \\
	conv3-1-2  &  29,393/36,864  & 20.27\% (1$\times$) &  4281/36,864  & 88.39\% (8$\times$) \\
	conv3-1-proj  &  1815/2048  & 11.38\% (1$\times$) &  638/2048  & 68.85\% (3$\times$) \\
	conv3-2-1  &  29,423/36,864  & 20.19\% (1$\times$) &  4399/36,864  & 88.07\% (8$\times$) \\
	conv3-2-2  &  29,372/36,864  & 20.32\% (1$\times$) &  4442/36,864  & 87.95\% (8$\times$) \\
	conv3-3-1  &  29,332/36,864  & 20.43\% (1$\times$) &  4427/36,864  & 87.99\% (8$\times$) \\
	conv3-3-2  &  29,244/36,864  & 20.67\% (1$\times$) &  4392/36,864  & 88.09\% (8$\times$) \\
	conv3-4-1  &  29,264/36,864  & 20.62\% (1$\times$) &  4450/36,864  & 87.93\% (8$\times$) \\
	conv3-4-2  &  28,954/36,864  & 21.46\% (1$\times$) &  4413/36,864  & 88.03\% (8$\times$) \\
	conv3-5-1  &  28,770/36,864  & 21.96\% (1$\times$) &  4469/36,864  & 87.88\% (8$\times$) \\
	conv3-5-2  &  28,455/36,864  & 22.81\% (1$\times$) &  4381/36,864  & 88.12\% (8$\times$) \\
	fc1  &  570/640  & 10.94\% (1$\times$) &  298/640  & 53.44\% (2$\times$) \\
	\midrule
	Total &  368452/464432 & 20.67\% (1$\times$) &  58051/464432 & 87.50\% (8$\times$) \\\midrule
	Test Accuracy & 0.9022 @ $\lambda$ = 1.001 & Ref: 0.9005 & 0.8922 @ $\lambda$ = 1.005 & Ref: 0.9005 \\\bottomrule
\end{tabular}}
\end{table}





\reftitle{References}






\end{document}